\documentclass[conference]{IEEEtran}
\IEEEoverridecommandlockouts
\usepackage{cite}
\usepackage{amsmath,amssymb,amsfonts}
\usepackage{algorithmic}
\usepackage{graphicx}
\usepackage{textcomp}
\usepackage{xcolor}
\def\BibTeX{{\rm B\kern-.05em{\sc i\kern-.025em b}\kern-.08em
    T\kern-.1667em\lower.7ex\hbox{E}\kern-.125emX}}
\begin{document}

\title{A Spatio-temporal Graph Network Allowing Incomplete Trajectory Input for Pedestrian Trajectory Prediction\\
}

\author{\IEEEauthorblockN{1\textsuperscript{st} Juncen Long}
\IEEEauthorblockA{
\textit{Politecnico di Milano}\\
Milan, Italy \\
juncen.long@polimi.it}
\and
\IEEEauthorblockN{2\textsuperscript{nd} Gianluca Bardaro}
\IEEEauthorblockA{\textit{Politecnico di Milano}\\
Milan, Italy \\
gianluca.bardaro@polimi.it}
\and
\IEEEauthorblockN{3\textsuperscript{rd} Simone Mentasti}
\IEEEauthorblockA{\textit{Politecnico di Milano}\\
Milan, Italy \\
simone.mentasti@polimi.it}
\and
\IEEEauthorblockN{4\textsuperscript{th} Matteo Matteucci}
\IEEEauthorblockA{\textit{Politecnico di Milano}\\
Milan, Italy \\
matteo.matteucci@polimi.it}
}

\maketitle

\begin{abstract}
Pedestrian trajectory prediction is important in the research of mobile robot navigation in environments with pedestrians. Most pedestrian trajectory prediction algorithms require the input historical trajectories to be complete. If a pedestrian is unobservable in any frame in the past, then its historical trajectory become incomplete, the algorithm will not predict its future trajectory. To address this limitation, we propose the STGN-IT, a spatio-temporal graph network allowing incomplete trajectory input, which can predict the future trajectories of pedestrians with incomplete historical trajectories. STGN-IT uses the spatio-temporal graph with an additional encoding method to represent the historical trajectories and observation states of pedestrians. Moreover, STGN-IT introduces static obstacles in the environment that may affect the future trajectories as nodes to further improve the prediction accuracy. A clustering algorithm is also applied in the construction of spatio-temporal graphs. Experiments on public datasets show that STGN-IT outperforms state of the art algorithms on these metrics.

\end{abstract}


\section{Introduction}

Many pedestrian trajectory prediction algorithms tried to help robots navigate in human-robot coexistence environments, from the early algorithms based on Bayesian filtering to the later algorithms based on recurrent neural networks (RNNs) and spatio-temporal graphs \cite{BF,SFM,SFMiP,SRLSTM,ARGB2}. Almost all existing algorithms use average displacement error (ADE) and final displacement error (FDE) as metrics. Furthermore, when a pedestrian is unobservable in any frame in the past, its historical trajectory is called as an incomplete trajectory, and these algorithms will not predict the future trajectory of this pedestrian.

\begin{figure}
  \begin{center}
  \includegraphics[width=3.5in]{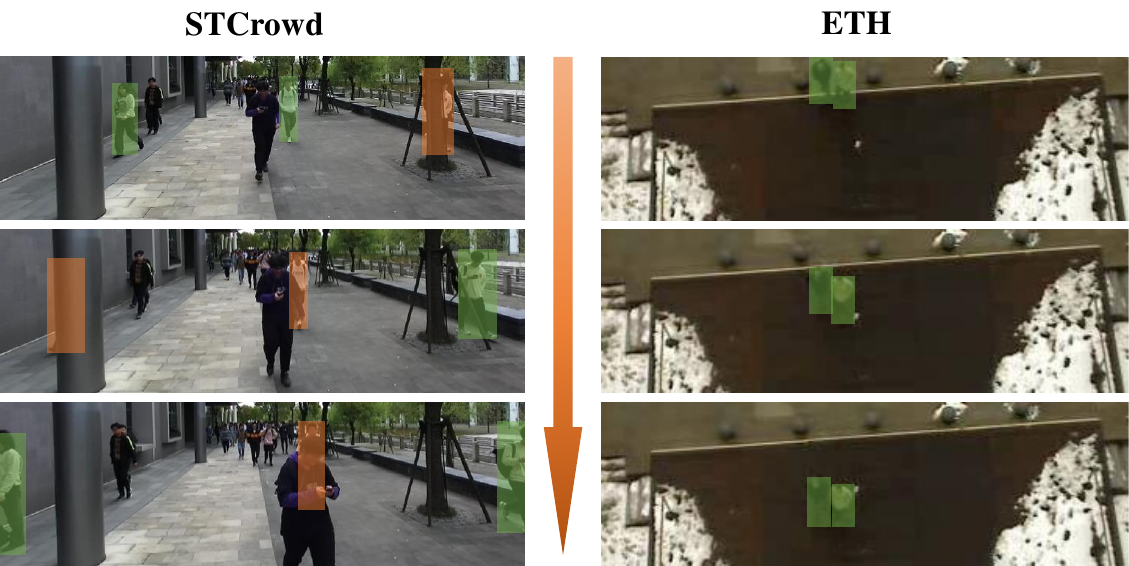}\\
  \caption{Comparison of the STCrowd dataset and the ETH dataset. Green boxes indicate pedestrians are observable and orange boxes indicate pedestrians are obscured. Pedestrians are more likely to be obscured in the egocentric view than in the top-down view.}
  \label{STC_ETH}
  \end{center}
\end{figure}

\begin{figure}
  \begin{center}
  \includegraphics[width=3.5in]{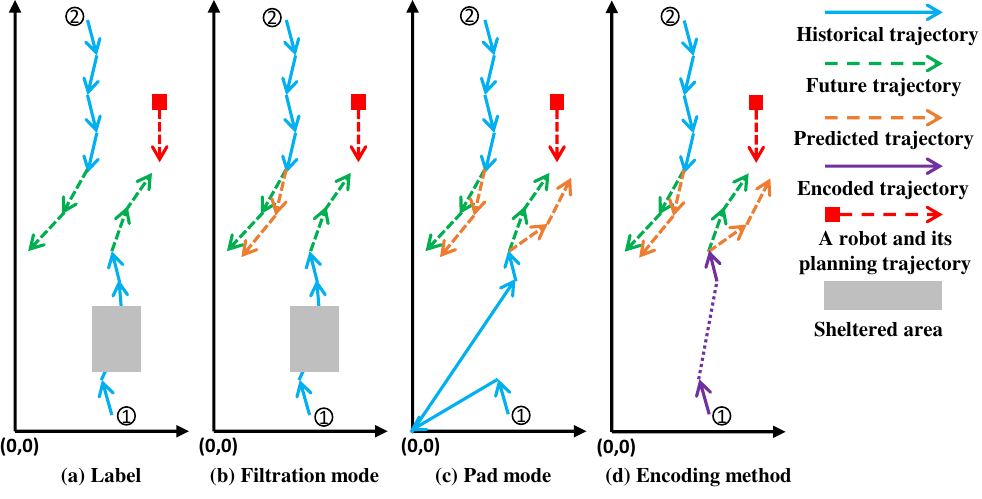}\\
  \caption{Label and Prediction results for filtration mode, pad mode, and encoding mode with incomplete trajectories. Incomplete trajectories are not predicted in filtration mode. Incomplete trajectories are predicted in pad mode with unobservable positions set to 0. The encoding method encodes the observation state of positions.}
  \label{FDE_FPA}
  \end{center}
\end{figure}

Although many pedestrian trajectory prediction algorithms have achieved excellent ADE and FDE, most of them are trained and evaluated on datasets with top-down views such as ETH \cite{ETH}, UCY \cite{UCY}, and SDD \cite{SDD}. However, the mobile robot usually obtains egocentric vision and local LIDAR maps, rather than a top-down view image. As shown in Fig.~\ref{STC_ETH}, pedestrians are unlikely to be obscured in the top-down view, but it is very common in the egocentric view, which means that incomplete trajectories have a greater impact on algorithms in egocentric view datasets than in top-down view datasets. 

As shown in Fig.~\ref{FDE_FPA}a, the historical trajectory of pedestrian 1 is incomplete, and the historical trajectory of pedestrian 2 is complete, while a robot may collide with pedestrian 1. The prediction mode of almost all existing algorithms, the filtration mode (Fig.~\ref{FDE_FPA}b), only predicts the trajectory of pedestrian 2. And the pad mode represents the pedestrian's position as (0,0) when it is obscured and then predicts its trajectory (Fig.~\ref{FDE_FPA}c). In this case, the FDE in filtration mode is less than that in pad mode, as it avoids the difficult prediction of incomplete trajectories. However, the pad mode is safer than the filtration mode as its prediction can indicate a possible collision for the robot. Therefore, it is more ideal to use pad mode and evaluate the performance of algorithms in pad mode.



For pad mode in Fig.~\ref{FDE_FPA}, it can be misinterpreted by algorithms that the pedestrian has moved from its original position to the position (0,0). Through experiments, we find that this misinterpretation can reduce the performance of algorithms.

\begin{figure}
  \begin{center}
  \includegraphics[width=3.5in]{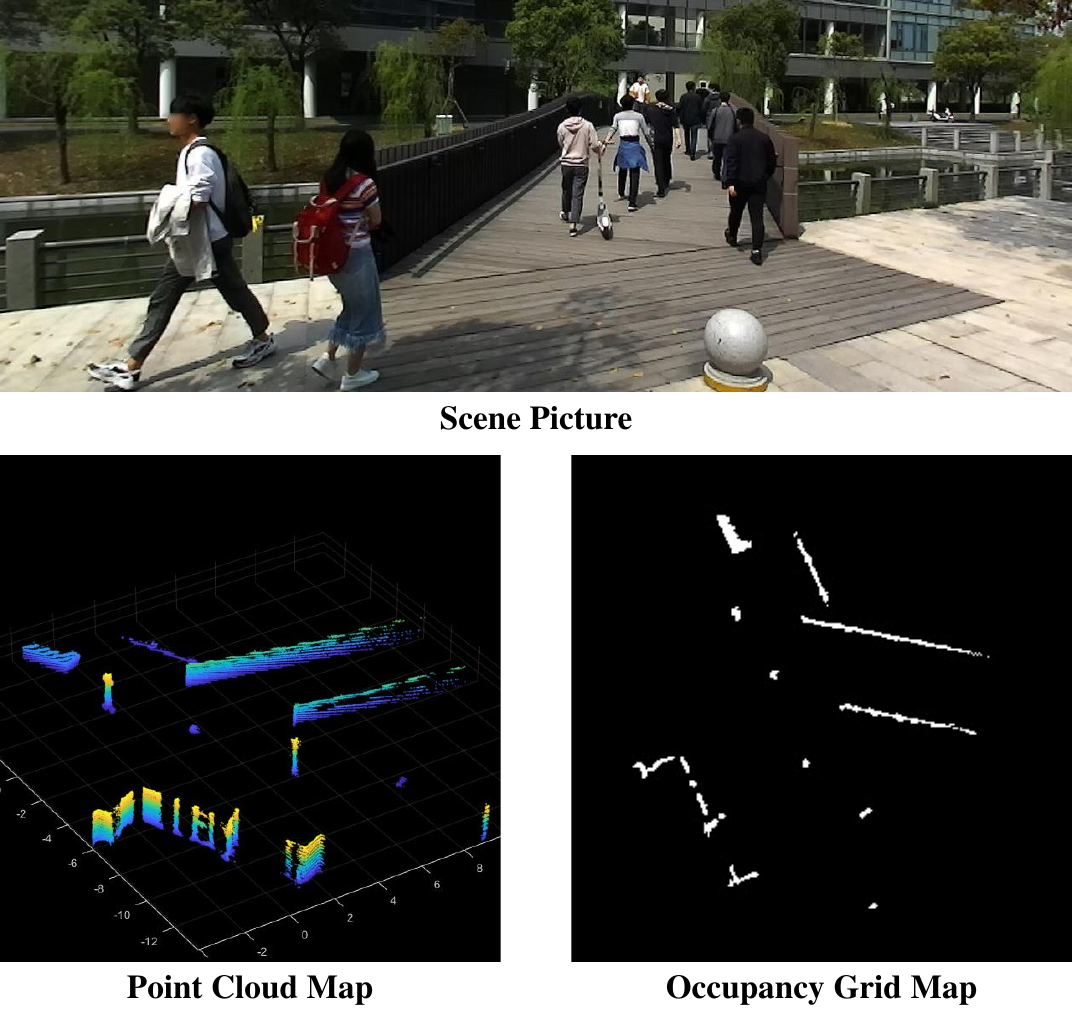}\\
  \caption{The scene picture (top), point cloud map (bottom left), and occupancy grid map (bottom right) of a scene in the STCrowd dataset. The occupancy grid map can be automatically generated from the point cloud map.}
  \label{map}
  \end{center}
\end{figure}

To address the limitations of existing algorithms in dealing with incomplete historical trajectories, we designed a spatio-temporal graph network allowing incomplete trajectory input (STGN-IT) to predict the future trajectories of pedestrians. As shown in Fig.~\ref{map}, STGN-IT obtains static obstacle information from the occupancy grid map, which can be automatically generated by the point cloud data. Thus, STGN-IT is more flexible than algorithms using semantic maps, which need to be manually labeled. In addition, STGN-IT uses spatio-temporal graphs to represent pedestrian and obstacle information and uses an encoding method to cope with the incomplete parts of historical trajectories. We evaluate our algorithm and compare it with state of the art algorithms on the public dataset STCrowd (STC) \cite{STCrowd}. 

The main contributions of this article are as follows.
\begin{enumerate}
\item	We designed STGN-IT, combining with a special encoding method, graph convolutional networks, and GRU networks. STGN-IT can predict the future trajectory of pedestrians with incomplete historical trajectories and occupancy grid maps, which makes it more suitable for robot navigation.


\item   We verify that the incomplete trajectories can seriously affect the performance of the existing algorithms by experiments, and propose an encoding method to reduce the performance degradation. 

\end{enumerate}

The paper is structured as follows; in Section 2, we present an overview of pedestrian trajectory prediction algorithms and datasets, focusing on available datasets and algorithms used for this task. In Section 3.A, we formulate
the problem. From Section 3.B, we present the STGN-IT algorithm with its pipeline for prediction in detail. In Section 4, we compare quantitatively and qualitatively the performance of different algorithms in different prediction modes with the ablation study.

\section{Related Works}

Datasets for pedestrian trajectory prediction have different perspectives. In top-down view datasets, ETH \cite{ETH} and UCY \cite{UCY} are used most frequently, and in recent years, SDD \cite{SDD} has become popular, which has more data and complex environments. In egocentric view datasets,  KITTI \cite{KITTI} and BDD100K use \cite{BDD100K} sensors carried on vehicles, while FPL \cite{FPL}, SiT \cite{SiT}, and STC \cite{STCrowd} use sensors carried by small robots or pedestrians. In the STC dataset, the sensors are carried on a static bracket, which makes it easier to model the impact on pedestrian trajectories. 

There are many existing algorithms based on long short-term memory (LSTM) networks \cite{LSTM} for trajectory prediction \cite{LSTMA1,LSTMA3,LSTMA4}. Given that the gated recurrent unit (GRU) \cite{GRU} has fewer parameters and similar performance to LSTM, some algorithms utilize GRU to predict trajectories \cite{GRUA1,GRUA2}. Some algorithms also use encoder-decoder structures to improve the performance \cite{en-de2}. Social-VRNN \cite{en-de3} and Social-BiGAT \cite{en-de1} use the encoder–decoder model and encode social interaction condition about pedestrians. A distribution discrimination method based on the encoder–decoder model, DisDis \cite{en-de4}, was proposed to learn the behavior pattern of each person.

Considering that interactions between pedestrians may occur at the end of the trajectories, some algorithms started to use the bidirectional LSTM and bidirectional GRU to extract trajectory features \cite{BiLSTMA1,BiLSTMA2,BiGRU1}. To better model interactions between pedestrians, some algorithms construct matrices using information of other pedestrians around a pedestrian, and extract features from matrices by neural network. The Social-LSTM \cite{SLS} is the most representative one, where the velocities and positions of the surrounding pedestrians are embedded within a 3D matrix by square segmentation. Except for that, SS-LSTM constructs the matrix using ring segmentation, and FSP constructs the matrix using relative positions \cite{FSP}.

Some algorithms use spatio-temporal graphs to represent historical trajectories of pedestrians. Typically, each node in the spatio-temporal graph represents a pedestrian, and the edges represent the correlations between pedestrians\cite{STGNE}. Social-STGCNN \cite{ST-CNN1}, a popular spatio-temporal graph based algorithm, uses relative velocity to represent edges. Then, features in the spatio-temporal graphs can be extracted by neural networks based on different structures \cite{ST-CNN2, M-SGCN}. For example, STAGP \cite{ST-CNN3} uses convolutional neural networks to extract features, GST \cite{ST-RNN1} uses recurrent neural networks, STAR \cite{ST-Tra1} uses transformers.

\section{Methodology}

\subsection{Problem Formulation}

The position information of pedestrian $i$ at time $t$ are represented as $X_t^i = [x_t^i,y_t^i]$. Further, $X_t^i = [x,y]$ when the pedestrian is observable with position $[x,y]$, and $X_t^i = [\text{Nan},\text{Nan}]$ when the pedestrian is not observable. Suppose there are $m$ pedestrians in the scene at time $t_0$, then their historical position information $H_{t_{0}}^{1:m}$ and ground-truth future position information $F_{t_{0}}^{1:m}$ can be represented as follows: 
\begin{equation} 
H_{t_{0}}^{i} = {\left \{{ {X_{t_{0}}^{i}, X_{t_{0}-1}^{i},\ldots,X_{t_{0}-T_{obs}+1}^{i}} }\right\}}
\label{PF_obs1}
\end{equation}

\begin{equation} 
H_{t_{0}}^{1:m} = {\left \{{ {H_{t_{0}}^{1}, H_{t_{0}}^{2},\ldots,H_{t_{0}}^{m}} }\right\}}.
\label{PF_obs2}
\end{equation}

\begin{equation} 
F_{t_{0}}^{i}  = {\left \{{ {X_{t_{0} +1}^{i}, X_{t_{0}+2}^{i}, \ldots,X_{t_{0} +T_{pred}}^{i}} }\right\}}
\label{PF_gt1}
\end{equation}

\begin{equation} 
F_{t_{0}}^{1:m} = {\left \{{ {F_{t_{0}}^{1} , F_{t_{0}}^{2} ,\ldots,F_{t_{0}}^{m} } }\right\}}.
\label{PF_gt2}
\end{equation}

\noindent
where $T_{obs}$ and $T_{pred}$ are the time step length of the historical and predicted trajectories, respectively. The algorithm will output future trajectory predictions $\hat {F}^{1:m}$ for $m$ pedestrians based on their historical trajectory information and environmental information. The loss function of STGN-IT tries to minimize the ADE between the prediction $\hat {F}^{1:m}$ and the ground-truth trajectory $F_{t_{0}}^{1:m} $.

\subsection{Algorithm Structure}
\begin{figure*}
  \begin{center}
  \includegraphics[width=6in]{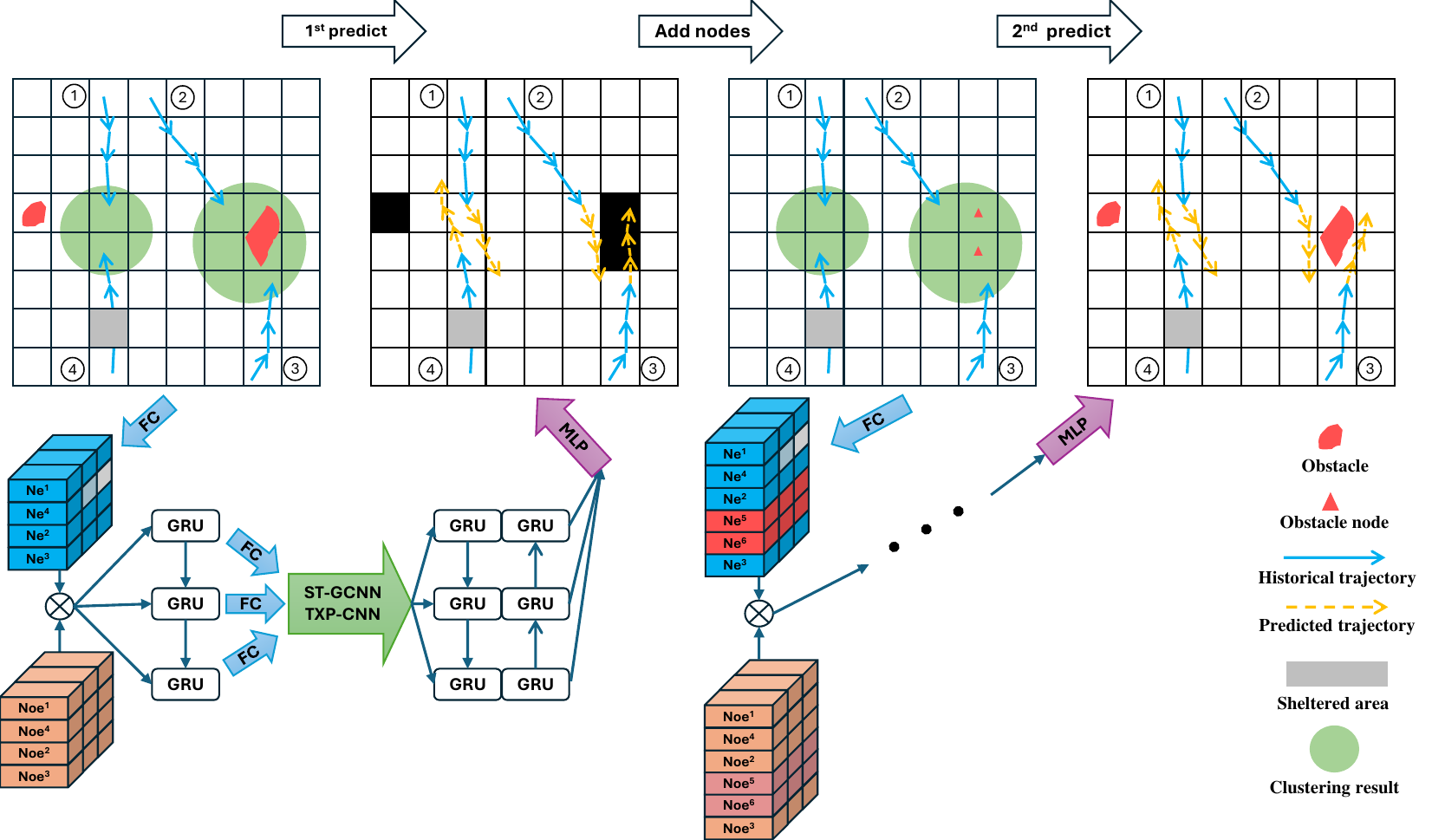}\\
  \caption{The STGN-IT algorithm includes two predictions. The first prediction searches for obstacles in the environment and adds them to the spatio-temporal graph to provide more information for the second prediction.}
  \label{structure}
  \end{center}
\end{figure*}

The STGN-IT algorithm makes 2 times of predictions and contains 4 modules. Its structure and prediction process are shown in Fig.~\ref{structure}. In the first prediction, the network directly predicts the future trajectories of pedestrians without using environmental information. Then obstacles near the predicted trajectory are added to the spatio-temporal graph as nodes, and the new spatio-temporal graph is used as input for the second prediction to increase the accuracy of the prediction. 

During the prediction, the spatio-temporal graph construction module generates the spatio-temporal graph and the corresponding matrices with the DBSCAN clustering algorithm \cite{DBSCAN}, then the observation state encoding module encodes the matrices based on the pedestrian observation states, and finally the trajectory prediction module predicts the trajectories based on the encoded features. Between two times of predictions, the obstacle addition module adds obstacles near the predicted trajectories to the graph as nodes.


\subsection{Spatio-temporal Graph Construction Module}

When constructing the spatio-temporal graph, we consider both position and velocity information. In the spatio-temporal graph, node $N_t^i$ represents the information of a pedestrian or an obstacle at time $t$, and edge $E_t^{ij}$ represents the correlation between $N_t^i$ and $N_t^j$ at time $t$. Suppose $\Delta X_t^i = X_t^i - X_{t-1}^i$ is the velocity of $N_t^i$ at time $t$, then $N_t^i$ and $E_t^{ij}$ can be represented as follows:

\begin{equation} 
N_t^i = [X_t^i,\Delta X_t^i]
\label{node_1}
\end{equation}

\begin{equation} 
E_t^{ij} = [X_t^i - X_t^j,\Delta X_t^i - \Delta X_t^j]
\label{edge_1}
\end{equation}

Then, we use the DBSCAN clustering algorithm to adjust the order of the nodes in the matrix, making the interactions between the nodes easier to detect. As shown in Fig.~\ref{structure}, the node order in the matrix is $(1,4,2,5,6,3)$ instead of $(1,2,3,4,5,6)$, because the obstacle $N^5$ and $N^6$ have an impact on the trajectories of the pedestrian $N^3$ and $N^2$, and making them neighboring in the matrix is beneficial for the graph convolutional network to extract the features.

\subsection{Observation State Encoding Module}

\begin{table}
\centering
\caption{Code Rules for $No_t^i$}
\label{No rule}
\renewcommand\arraystretch{1.2}
\begin{tabular}{| c c |c|}
\hline
$ON_t^i$ & 
$ON_{t-1}^i$ & 
$No_t^i$ \\
\hline
True		& True			& [1,1,1,1]\\
True		& False			& [1,1,0,0]\\
False		& /			    & [0,0,0,0]\\

\hline
\end{tabular}
\end{table}

\begin{table}
\centering
\caption{Code Rules for $Eo_t^i$}
\label{Eo rule}
\renewcommand\arraystretch{1.2}
\begin{tabular}{|c c |c|}
\hline
$ON_t^i$ \& $ON_t^j$ & 
$ON_{t-1}^i$ \& $ON_{t-1}^j$ & 
$Eo_t^{ij}$ \\
\hline
True		& True			& [1,1,1,1]\\
True		& False			& [1,1,0,0]\\
False		& /			    & [0,0,0,0]\\

\hline
\end{tabular}
\end{table}

When node $N_t^i$ is not observable, we let $X_t^i = [0,0]$ and $\Delta X_t^i = \Delta X_{t+1}^i = [0,0]$. To allow the network to distinguish whether a node is not observable or truly in position (0,0), we design an encoding rule to describe the observation state. Specifically, we add two vectors, $No_t^i$ and $Eo_t^i$, to describe the availability of the data. Define $ON_t^i$ as the observation state of $N_t^i$, the specific rules are shown in Table~\ref{No rule} and Table~\ref{Eo rule}.

Then, for the nodes, we use the fully connected layers to combine the information from $N_t^i$ and $No_t^i$ to obtain the feature $Nf_t^i$:
\begin{equation} 
Ne_t^i = \phi^{ne}(N_t^i;W_{ne}) 
\label{EN}
\end{equation}

\begin{equation} 
Noe_t^i = \phi^{noe}(No_t^i;W_{noe}) 
\label{ENo}
\end{equation}

\begin{equation} 
Nf_t^i = {Ne_t^i} \odot {Noe_t^i} 
\label{Nf}
\end{equation}

\noindent
where $\phi^{ne}$ and $\phi^{noe}$ are the fully connected layers with an input dimension of 4 and an output dimension of $n_{en}$, and $\odot$ represents the Hadamard product. 

By using another two fully connected layers $\phi^{ee}$ and $\phi^{eoe}$, and a similar process to (\ref{EN})-(\ref{Nf}), we can also obtain the edge feature $Ef_t^{ij}$. Then, by embedding them into the corresponding places of the matrices, the new node matrix $Vf$ and the adjacency matrix $Af$ can be constructed. The dimension of $Vf$ is $[T_{obs},m,n_{en}]$ and the dimension of $Af$ is $[T_{obs},m,m,n_{en}]$.

\subsection{Trajectory Prediction Module}

In order to reduce the influence of missing positions on the trajectory feature extraction, two GRU networks are first used to compensate for the missing position information by utilizing the features from previous frames. The compensation node matrix $Vfc$ can be constructed as follows:

\begin{equation} 
Hvfc = {\text{GRU}}^{vf}(Hvf, Vf ;W_{vf})   
\label{GRU_HDV}
\end{equation}

\begin{equation} 
Vfc = \phi ^{vfc}(Hvfc;W_{vfc})
\label{FC_HDV}
\end{equation}

\noindent
where ${\text{GRU}}^{vf}$ is a GRU network with an input layer dimension of $n_{en}$ and hidden state dimension of $n_{gru}$, and $\phi ^{vfc}$ is the fully connected layer with an input dimension of $Ngru$ and an output dimension of $n_{en}$. $Hvf$ is the initial hidden state of $\text{BiGRU}^{vf}$, and $Hvfc$ is the stack of hidden states at each step of $\text{BiGRU}^{vf}$. Similarly, we construct the compensation adjacency matrix $Afc$ by another two networks ${\text{BiGRU}}^{af}$ and $\phi^{afc}$, and a similar process to (\ref{GRU_HDV})-(\ref{FC_HDV}). $Vfc$ has the same dimension as $Vf$, and $Afc$ has the same dimension as $Af$.

Inspired by \cite{ST-CNN1}, we use the Spatio-Temporal Graph Convolution Network (STGCN) and the Time-Extrapolator Convolution Network (TECN) to extract features from the $Vfc$ and $Afc$ matrices. The process is shown as follows:

\begin{equation} 
Vstg = {\text{STGCN}}(Vfc, Afc ;W_{stgcn})  
\label{STGCN}
\end{equation}

\begin{equation} 
Vp = {\text{TECN}}(Vstg ;W_{tecn})  
\label{TECN}
\end{equation}

\noindent
where ${\text{STGCN}}$ is a STGCN network that kernel size is $n_{stg}$, and ${\text{TECN}}$ is a three-layer TECN Network that kernel size is $n_{te}$. The node prediction matrix $Vp$ has a dimension of $[T_{pred},m,n_{de}]$. 

Finally, a Bi-GRU network is utilized to extract the features in $Vp$ to obtain the matrix $GVp$, and then a multi-layer perception (MLP) network is utilized to decode $GVp$ and output the final prediction velocity. The process is shown as follows:

\begin{equation} 
Sp_{1:T_{pred}} = {\text{BiGRU}}^{gvp}(Hgvp, GVp ;W_{pv})   
\label{GRU_PDV}
\end{equation}

\begin{equation} 
Dx_t = {\text{MLP}}^{sp}(Sp_t;W_{sp}) \quad {(t=1,2,\ldots,T_{pred})}  
\label{MLP_PDV}
\end{equation}

\begin{equation} 
\hat X_t = \hat X_{t-1} + Dx_t \quad {(t=1,2,\ldots,T_{pred})}  
\label{RX}
\end{equation}

\noindent
where ${\text{BiGRU}}^{gvp}$ is a Bi-GRU network with an input layer dimension of $n_{de}$ and hidden state dimension of $n_{gru}$, and $\text{MLP}^{sp}$ is a three-layer MLP network with an input dimension of $2 * n_{gru}$ and an output dimension of 2. $Hgvp$ is the initial hidden state of $\text{BiGRU}^{gvp}$, and $Sp_{1:T_{pred}}$ is the stack of hidden states at each step of $\text{BiGRU}^{gvp}$. $Dx_t$ and $\hat X_t$ are the displacement and position of pedestrians predicted by the network at time $t$, respectively.

\subsection{Obstacle Addition Module}

As shown in Fig.~\ref{structure}, after the first prediction, the obstacle addition module will add obstacle nodes to the spatio-temporal graph based on the occupancy grid map and the predicted trajectories. Suppose the set of predicted trajectories in the first prediction is $\hat {\boldsymbol{X}} $. The obstacle position in the occupancy grid map can be represented as $(x_{obs}^i,y_{obs}^i)$, and the set of obstacles added to the spatio-temporal graph is $\boldsymbol{Obs}$, defined as follows:

\begin{equation} 
\boldsymbol{Obs} = \left \{ (x_{obs},y_{obs}) | f_{\text{mindis}}((x_{obs},y_{obs}),\hat {\boldsymbol{X}}) < od \right \}  
\label{mindis}
\end{equation} 

\noindent
where the function $f_{\text{mindis}}(p,S)$ is defined as the minimum distance from the point $p$ to all points in the set $S$, and $od$ is the minimum distance to add an obstacle as a node. 

Since the obstacles in $\boldsymbol{Obs}$ are close to the pedestrian's predicted trajectories, they will affect the future trajectories of pedestrians. After adding them to the spatio-temporal graph as nodes, STGN-IT will make the second prediction, which considers the influence of the environment and has a higher accuracy.

\section{Experiments and Analysis}

\subsection{Evaluation Metrics}
We use average displacement error (ADE) and final displacement error (FDE) to evaluate the performance of algorithms, which are defined as follows:

\begin{equation} 
\text{ADE} = \frac{\sum _{i=1}^m {\sum _{t=1}^{T_{\text{pred}}} \left\|{ \hat X_t^i -X_t^i}\right\|_2 }}{m*T_{\text{pred}}}
\label{ADE}
\end{equation}

\begin{equation} 
\text{FDE} = \frac{\sum _{i=1}^m \left\|{ \hat X_{T_{\text{pred}}}^i -X_{T_{\text{pred}}}^i}\right\|_2 }{m}
\label{FDE}
\end{equation}

\noindent
where $\left\|{\cdot} \right\|_2$ denotes the Euclidean norm. ADE and FDE are measured in meters.

\subsection{Dataset and Parameter Settings}

The STC dataset annotated over 5,000 trajectories in over 10 scenes with the raw data of 3D LIDAR for each scene. We use the raw LIDAR data to create point cloud maps and further generate occupancy grid maps. We use trajectories in the STC dataset to train and evaluate the performance of STGN-IT and state of the art algorithms. To further evaluate the influence of missing trajectories on the algorithm performance, we randomly removed about 10\% of the samples from the original dataset and generated a new dataset, STC-c. Samples are kept when they are used as labels and only removed when they are input as observation sequences, so STC and STC-c datasets have the same training and validation labels. 

The STC dataset has a frame rate of 2.5 Hz. Same as the settings of state of the art algorithms, we set the observation time as 3.2 seconds and the prediction time as 4.8 seconds, so $T_{obs}=8$ and $T_{pred}=12$. As defined in (\ref{condi}), the conditions for STGN-IT to perform trajectory prediction for pedestrian $i$ are that it is observable in the latest frame and is observable for over 2 of the past 8 frames. Notice that under this rule, STGN-IT predicts the trajectory of a pedestrian 1.2 seconds after observing it, rather than the 3.2 seconds required by most existing algorithms. The shorter response time makes STGN-IT more suitable for mobile robot navigation than other existing algorithms.

\begin{equation} 
 {X_0^i \neq [\text{Nan},\text{Nan}]} \ \&     
 {\sum _{t=-T_{obs}+1}^{0} f_\text{exist} (X_t^i) > 2 }
\label{condi}
\end{equation}

\begin{equation} 
f_{\text{exist}}(X)=\begin{cases}
    1 & \text{if} \ X \neq [\text{Nan},\text{Nan}] \\
    0 & \text{if} \ X = [\text{Nan},\text{Nan}]
\end{cases}
\label{fexist}
\end{equation}

The structural parameters of the network are set as $n_{en}=9$, $n_{de}=7$, $n_{gru}=64$, $n_{stg}=7$, $n_{te} = 3$. The distance parameters are set as $od=0.8$, $ad=1$, $fd=1$. The learning rate is set to 0.001, the batch size is set to 16, and the number of epochs is set to 200.

\subsection{Experiment Settings}

\begin{figure*}
  \begin{center}
  \includegraphics[width=5.5in]{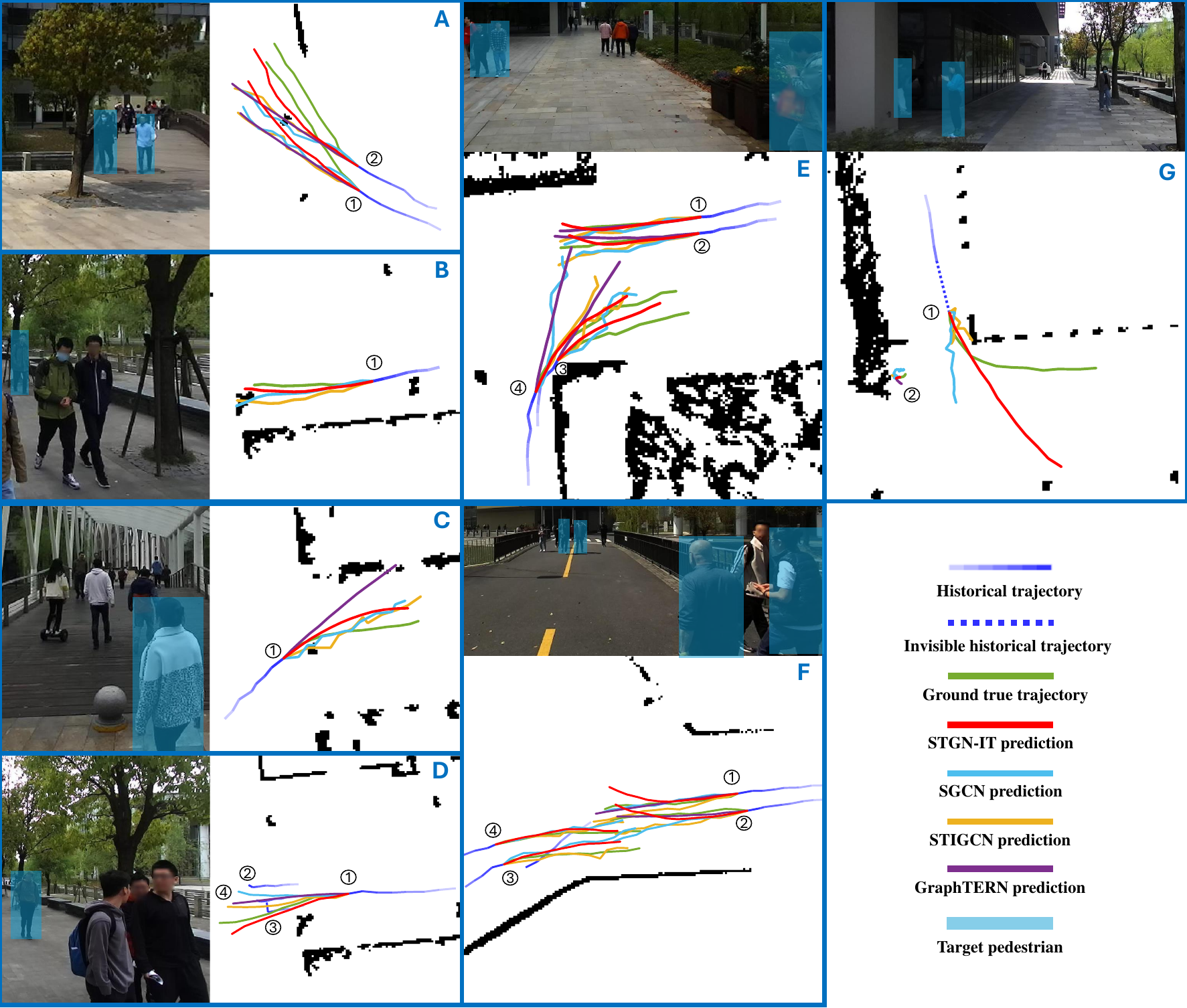}\\
  \caption{The prediction results of STGN-IT and some state of the art algorithms. GraphTERN makes predictions in condition ``f-f''. STGN-IT, SGCN, STIGCN make predictions in condition ``p-p''. GraphTERN does not make predictions in scene B, F, G, indicating that the ``f-f'' prediction condition is more likely to cause robot collisions. The trajectories predicted by the STGN-IT are more reasonable, as the trajectories avoid static obstacles in scenes A, B, C and avoid other pedestrians in scenes D, E, F. With incomplete trajectory input in scene G, predictions of STGN-IT are smoother and more reasonable.}
  \label{visi}
  \end{center}
\end{figure*}

We compare STGN-IT with the following state of the art algorithms: STIGCN\cite{STIGCN} (2024), SSAGCN\cite{ARGB1} (2023), MSRL\cite{MSRL} (2023), GraphTERN\cite{GTERN} (2023), Social-Implicit\cite{SoicalI} (2022), SGCN\cite{SGCN} (2021), Social-STGCNN\cite{ST-CNN1} (2020).

Referring to \cite{bestk3}, for the state of the art algorithms, we randomly sample 3 times and select the samples with the best metrics, that is, $\text{minADE}_{3}$ and $\text{minFDE}_{3}$. For STGN-IT, we output three candidate trajectories based on the possible directions of pedestrians. To better evaluate the performance of the algorithms, we train and evaluate them with two modes, the filtration mode and the pad mode:
\begin{itemize}
\item	Filtration mode: the algorithms only predict pedestrians with complete historical trajectories.
\item	Pad mode: the algorithms predict pedestrians that satisfy (\ref{condi}) during training and testing. When pedestrians are not observable, their positions are replaced by (0,0).
\end{itemize}

In the following, we refer to the condition ``p-p'' as the algorithm trained and tested with filtration mode, and refer to the condition ``p-f'' as the algorithm trained with pad mode and tested with filtration mode.

\subsection{Quantitative Experiments and Analysis}


\begin{table}
\centering
\caption{The $\text{minADE}_{3}$/$\text{minFDE}_{3}$ results, the lower number indicates better performance. \textbf{Bold} and \underline{underlined} mark the best and second-best results.}
\label{result1}

\begin{tabular}{|c|c|c|c|}
\hline
& 
STC,f-f $\downarrow$ &
STC,p-p $\downarrow$ & 
STC-c,p-p $\downarrow$ \\
\hline
Social-STGCNN    & 0.68/1.18 & 0.74/1.32 & 0.84/1.49 \\
SGCN		     & 0.38/0.69 & \underline{0.43}/\underline{0.77} & 0.67/1.19 \\
Social-Implicit	 & 0.38/0.76 & 0.55/1.06 & 0.80/1.50 \\
GraphTERN	     & \underline{0.34}/\underline{0.66} & Nan/Nan   & Nan/Nan   \\
MSRL	         & 0.41/0.77 & 0.58/0.92 & 0.69/0.98 \\
SSAGCN	         & 0.41/0.74 & 0.53/0.96 & 0.78/1.40 \\
STIGCN	         & 0.39/0.71 & 0.46/0.84 & \underline{0.51}/\underline{0.89} \\
STGN-IT	         & \textbf{0.30}/\textbf{0.56} & \textbf{0.35}/\textbf{0.62} & \textbf{0.35}/\textbf{0.64}\\

\hline
\end{tabular}
\end{table}

Table~\ref{result1} shows the ADE and FDE of algorithms evaluated in three prediction conditions. The performance of all algorithms decreases from ``STC,f-f'' to ``STC,p-p'' to ``STCc,p-p'', due to the increasing incomplete parts of the trajectories between these three conditions. However, the performance degradation rate differs greatly among different algorithms. GraphTERN cannot be trained successfully in condition ``p-p'', SSAGCN and Social-Implici have almost twice the metrics in ``STCc,p-p'' as `STC,f-f'', and even metrics for algorithms that are less affected, such as STIGCN and MSRL, also increase by over 25\%. This situation leads to the second best algorithm being different in three prediction conditions. STGN-IT, on the other hand, has a performance degradation rate of about 15\%, which is the smallest among all the algorithms. Also, STGN-IT has the best ADE and FDE in all three conditions, which means that the predictions of STGN-IT are closest to the true trajectories of pedestrians, making potential collisions the most likely to be detected.

Note that the ADE and FDE in condition ``f-f'' are the metrics used in most publications, however, as demonstrated in Fig.~\ref{STC_ETH} and Fig.~\ref{FDE_FPA}, condition ``p-p'' is safer for robot navigation rather than condition ``f-f''. And in Table~\ref{result1}, an algorithm has a low ADE in condition ``f-f'' does not mean a low ADE in condition ``p-p'', as Social-Implicit has lower ADE than STIGCN in condition ``f-f'' but higher ADE than STIGCN in condition ``p-p''. Therefore, when evaluating the performance of algorithms, it is more appropriate to compute the ADE in condition ``p-p'' rather than in condition ``f-f''.

\begin{table}
\centering
\caption{The $\text{minADE}_{3}$/$\text{minFDE}_{3}$ results of ablation study. \textbf{Bold} and \underline{underlined} mark the best and second-best results.}
\label{result3}
\begin{tabular}{|c|c|c|}
\hline
& 
STC,f-f $\downarrow$ &
STC,p-p $\downarrow$  \\
\hline
STGN-IT w/o obs  & 0.41/0.78 &\underline{0.43}/\underline{0.79} \\
STGN-IT w/o code & \underline{0.32/0.58} & 0.46/0.85 \\
STGN-IT w/o clu	 & 0.40/0.74 & 0.48/0.93 \\
STGN-IT	         &\textbf{0.30}/\textbf{0.56} & \textbf{0.35}/\textbf{0.62}  \\
\hline
\end{tabular}
\end{table}

\subsection{Ablation Study}
We explore the influence of different modules on the performance of STGN-IT through an ablation experiment with the following algorithms:
\begin{enumerate}
\item	STGN-IT without adding obstacle nodes. (w/o obs)
\item	STGN-IT without observation state. (w/o code)
\item	STGN-IT without clustering process. (w/o clu)
\end{enumerate}

Table~\ref{result3} shows the ADE and FDE of algorithms evaluated in different prediction conditions. The least affected algorithm is STGN-IT w/o code in condition ``f-f'', which makes sense because in condition ``f-f'' all input trajectories are complete and the observation state encoding is redundant. In addition to this case, the deletion of any module reduces the performance metrics of the algorithm by at least 20\%. 

Without the clustering process, the feature vectors of nodes with interactions cannot be neighboring in the matrices, making the network hard to extract features. Without the observation state encoding, the algorithm cannot recognize invisible trajectories properly, thus reducing performance. Without adding obstacle information, the algorithm cannot predict the trajectories of pedestrians to avoid obstacles, resulting in lower performance.

\subsection{Qualitative Analysis}

The prediction results of some scenes are shown in Fig.~\ref{visi}. In scenes B, F, and G, GraphTERN does not predict the future trajectories of some pedestrians due to their incomplete historical trajectories, which can increase the risk of pedestrians during robot navigation. In scenes A, B, and C, the trajectories predicted by some state of the art algorithms clash with static obstacles, while the predictions of STGN-IT do not. This is because STGN-IT uses the occupancy grid map as the input, reducing the probability of predicting collision trajectories. 

STGN-IT also successfully predicts the interactions between pedestrians. In scene D, pedestrians 2, 3, and 4 stop on the road, and pedestrian 1 bypasses them, and STGN-IT successfully predicts the bypass trajectory. In scene E, four pedestrians meet at an intersection, and the STGN-IT successfully predicts their turns, while the trajectories predicted by the SGCN collide. The same situation occurs in scene F, where only the trajectories predicted by STGN-IT avoid collisions, GraphTERN does not predict trajectories, and the trajectories predicted by SGCN and STIGCN collide. 

In scene G, when the trajectory of pedestrian 1 is partially missing due to column occlusion, the trajectories predicted by STGN-IT are smooth and roughly correct, while the trajectories predicted by STIGCN and SGCN are very unstable. The trajectory predictions for Pedestrian 2 demonstrate that all algorithms have good predictions for stationary pedestrians.

Qualitative analysis demonstrates that STGN-IT has good trajectory prediction performance and can predict trajectories that are smooth and close to the ground-truth labels.

\section{Conclusion}

In this paper, we present a spatio-temporal graph network allowing incomplete trajectory input (STGN-IT) for pedestrian trajectory prediction. By using occupancy grid maps, observation state encoding, and clustering processes, STGN-IT achieves better trajectory prediction accuracy than the other algorithms in experiments. In addition, we propose the prediction condition of the pad mode, which is more ideal for applying the algorithm to mobile robot navigation than the filtration mode. For incomplete trajectory inputs in the pad mode, STGN-IT can output predictions more consistently than other state of the art algorithms.

Based on STGN-IT, we will further study algorithms to improve the prediction accuracy of pedestrian trajectory prediction with incomplete historical trajectories. 

\bibliographystyle{IEEEtran}
\bibliography{IEEEabrv, Bibliography.bib}

\end{document}